# Inferring Implicit 3D Representations from Human Figures on Pictorial Maps

Raimund Schnürer[a]*, A. Cengiz Öztireli[b], Magnus Heitzler[a], René Sieber[a], Lorenz Hurni[a]

[a] *Department of Civil, Environmental and Geomatic Engineering, ETH Zurich, Zurich, Switzerland;*

[b] *Department of Computer Science and Technology, University of Cambridge, United Kingdom*

* Institute of Cartography and Geoinformation (HIL G 14.2), Stefano-Franscini-Platz 5, 8093 Zurich, Switzerland; schnuerer@ethz.ch

# Inferring Implicit 3D Representations from Human Figures on Pictorial Maps

In this work, we present an automated workflow to bring human figures, one of the most frequently appearing entities on pictorial maps, to the third dimension. Our workflow is based on training data and neural networks for single-view 3D reconstruction of real humans from photos. We first let a network consisting of fully connected layers estimate the depth coordinate of 2D pose points. The gained 3D pose points are inputted together with 2D masks of body parts into a deep implicit surface network to infer 3D signed distance fields (SDFs). By assembling all body parts, we derive 2D depth images and body part masks of the whole figure for different views, which are fed into a fully convolutional network to predict UV images. These UV images and the texture for the given perspective are inserted into a generative network to inpaint the textures for the other views. The textures are enhanced by a cartoonization network and facial details are resynthesized by an autoencoder. Finally, the generated textures are assigned to the inferred body parts in a ray marcher. We test our workflow with 12 pictorial human figures after having validated several network configurations. The created 3D models look generally promising, especially when considering the challenges of silhouette-based 3D recovery and real-time rendering of the implicit SDFs. Further improvement is needed to reduce gaps between the body parts and to add pictorial details to the textures. Overall, the constructed figures may be used for animation and storytelling in digital 3D maps.

# 1. Introduction

Technology companies - such as Meta, Microsoft or Sony - invest heavily in the creation of the metaverse these days (Gilbert, 2021). In the visions of these companies (e.g., Meta, 2021), people equipped with head-mounted displays can immerse in virtual 3D environments for work or leisure activities. The virtual space may represent our physical world, be purely imaginary or be a mixture of both. Creating digital 3D representations of topographic elements and thematic content from the real world by abstraction and generalization would be of interest from a cartographic perspective. Early works have focused the rendering of sketched 3D buildings (Döllner & Walther, 2003) or the modelling of pictorial 3D mountains and sights (Naz, 2005). This is opposed to 'mirror worlds', for instance in Google Earth, where the real world is convincingly reflected (Park & Kim, 2022) and photo-realistically rendered. Cartographic 3D models and mirror worlds are closely related to 'digital twins', which are virtual representations of real world entities for mainly simulation purposes (Park & Kim, 2022), for example historical reconstructions (Herold & Hecht, 2018) or urban planning (Schrotter & Hürzeler, 2020).

One key concept of the metaverse are avatars (Park & Kim, 2022). Those virtual 3D models of humans, animals, or other personifications embody real humans or computer-controlled entities, who can be interacted with. In a cartographic 3D environment, avatars may give background information to a topic, tell personal stories or serve as tour guides. For example, 3D figures could illustrate daily life (e.g., farmer on a field), act in special events (e.g., priest at a coronation ceremony), or represent famous persons (e.g., Goethe in Weimar). Past research has examined the animation of 3D objects, such as cars and horses, on the terrain (Evangelidis et al., 2018) and the integration of 3D characters into cartographic virtual reality environments (Matthys et al., 2021).

The enrichment of cartographic digital twins with human figures would be an analogy to historic maps, where human figures have been inserted for ethnographic or humoristic purposes, amongst others (Child, 1956). Historic, but also contemporary pictorial maps would be valuable sources for creating 3D models of the depicted figures. Similarly to cartoons (X. Wang & Yu, 2020), pictorial figures on maps are usually composed of rather geometrically formed and possibly disproportionate shapes, whose low-detail textures are filled by flat colors and accentuated by sharp black edges. Nevertheless, the manual creation of 3D figures would be labor-intensive and cumbersome, and parametric models may not be detailed enough. A promising technique to reconstruct 3D models from humans and objects on photos are machine learning methods (Fahim et al., 2021). In cartography, researchers rather focused on the detection of topographic elements on maps – such as buildings (Heitzler & Hurni, 2020) or water bodies (Wu et al., 2022) – or pictorial figures (Schnürer, Öztireli, et al., 2021) by convolutional neural networks (CNNs) so far, however not with their transfer into the 3D space yet.

In this work, we like to close this gap by applying a series of neural networks to infer 3D representations, encoded as signed distance fields (SDFs), from 2D figures on pictorial maps. Each point in an SDF holds a value denoting the distance to the nearest boundary of an object. As a difference to previous works, we do not recover the SDFs from textures but merely from silhouettes. Compared to meshes, point clouds, or voxels, implicit representations like SDFs have some desirable properties such as an infinite geometric detail or easy blending capabilities. We use sphere tracing (Hart, 1996) to render the figures in real-time, whereas other researchers mainly used marching cubes

(Lorensen & Cline, 1987) to polygonise the SDFs. The sphere tracing algorithm is relatively well-established in contrast to newer methods like neural rendering (e.g., Eslami et al., 2018; Lassner & Zollhöfer, 2021). Our work has a great potential for skeletal animation since we construct the figures by compositing 3D body parts according to 3D pose points. We see atlases, education, museums, tourism or games in- and outside the metaverse as primary application areas.

## 2. Related work

In recent years, many advances have been made in reconstructing 3D persons and objects from single images using machine learning methods. For example, Omran et al. (2018) apply CNNs to predict parameters of pose and shape of a 3D person model by taking advantage of segmented body parts as an intermediate representation. Saito et al. (2019) developed the 'PIFu' architecture, which produces a 3D occupancy field for the geometry of a person by a multilayer perceptron (MLP) and texture colors by a generative adversarial network (GAN). In 'ARCH', described by Huang et al. (2020), animation capabilities of human models are considered by including a semantic deformation field, amongst others. Lin et al. (2020) encode images of objects in a hypernetwork, which is a network generating weights for another network. In the architecture of Lin et al. (2020), the hypernetwork predicts parameters of implicit functions for an MLP, which converts encoded 3D coordinates into SDF and RGB values and which is updated by a recurrent neural network.

In a subset of single view 3D reconstruction networks, coarse and detailed geometry are handled separately. In the deep implicit surface network (DISN), proposed by Wang et al. (2019), SDFs are predicted from local and global features, which are extracted from feature maps of an image encoder. Branches for coarse and fine level geometry exist also in 'PifuHD' (Saito et al., 2020). The successor of 'PIFu' contains two CNNs, three MLPs and a conditional GAN predicting normal maps. Li & Zhang (2021) demonstrated with 'D²IM-Net' how to transfer surface details from displacement maps to coarse shapes by one image encoder and two decoder branches as well as including a Laplacian loss function.

While the above networks use 3D training data, another subgroup of object reconstruction networks, also known as off-the-shelf recognition systems, uses only the given 2D images for supervision. Liu et al. (2019) elaborated a ray-based field probing technique to correct errors of predicted 3D implicit surfaces. Lunz et al. (2020) added a proxy neural renderer to a GAN to render 2D images by the traditional non-differentiable rendering pipeline. In 'U-CMR', Goel et al. (2020) optimized possible camera views to render meshes and textures of objects and birds. Ye (2021) render images from semi-implicit volumetric representations and only take approximate instance segmentation masks into account for supervision. In 'pixelNeRF' (Yu et al., 2021), volumetric density and color of objects are implicitly encoded along camera rays by a CNN.

A third subcategory of networks additionally outputs distinct parts or part memberships for the 3D reconstruction. In an early work, Agarwal and Triggs (2006) approximated body parts by cuboids using non-linear regression with Support Vector Machines. In a more modern architecture, Niu et al. (2018) extracted object parts as cuboids by sequential CNNs recovering masks and hierarchies. Paschalidou et al. (2020) trained a partition network to split objects into two parts, a geometry network to find shape parameters of geometric primitives, and a structure network which links the partitions to the primitives.

Instead of reconstructing the object out of individual shapes, Varol et al. (2018) relied in 'BodyNet' on a voxel-based representation, which is predicted by CNNs together with estimating 2D and 3D poses and segmenting 2D body parts. A more fine-grained part membership than a body part segmentation are UV-coordinates, which link texture images to the surface of a 3D model. UV-coordinates can be predicted from images and also be used for 3D reconstruction (e.g., Güler et al., 2017; Yao et al., 2019).

A last group of networks related to our research is concerned with the reconstruction of objects and figures based on silhouettes or sketches. Di and Yu (2017) propose a stacked hierarchical network consisting of 3D CNNs to create objects from black-and-white silhouette images. Delanoy et al. (2018) reconstruct voxelized objects from sketches by an image encoder-decoder CNN and a updater CNN. Brodt & Bessmeltsev (2022) recover 3D poses from sketched characters by training a 2D pose estimation network and applying an optimization algorithm focusing on bone tangents, body part contacts and bone foreshortening. No literature could be found to reconstruct 3D persons or objects from paintings, comics/manga, or maps by machine learning methods.

In this work, we address this shortage by following a bottom-up approach, distantly related to deep local shapes (Chabra et al., 2020), to build pictorial human figures from individual body parts. In a top-down approach, contrariwise, it may be more challenging to identify 3D body parts after having constructed a holistic 3D model. The enclosure of 3D body parts into bounding boxes may accelerate sphere tracing computations and may lead to less storage compared to covering the full 3D space. As the variance of SDF values within the bounding boxes is lower than the variance of the entire body, a more efficient training process and more fine-grained reconstruction results can be expected. For deriving 3D body parts from their 2D silhouettes, we use the DISN architecture (W. Wang et al., 2019) due to its simplicity and adaptability. 3D skeletal points, whose depth coordinates are predicted by another minimalistic network (Martinez et al., 2017), serve as anchor points for creating the 3D body parts. Textures based on the given view are generated by an inpainting network (Grigorev et al., 2019) using UV-coordinates predicted by a U-Net (Ronneberger et al., 2015). Finally, the textures are enhanced by a cartoonization network (X. Wang & Yu,

2020) and an autoencoder (Gondara, 2016). Overall, we aim at providing an easily understandable yet effective pipeline for inferring implicit 3D representations of pictorial figures.

## 3. Data

Generalized 3D body meshes of a female and male person from the SMPL-X dataset (Pavlakos et al., 2019) form the basis for our experiments. In the following, we process the meshes (Figure 1) by a Blender plugin provided for the SMPL-X dataset and automate the steps with the Blender scripting API. We assign about 3200 poses from the AGORA dataset (Patel et al., 2021) to the meshes, half to females and the other half to males. Additionally, we vary height and weight parameters (i.e., 1.40m & 60kg, 1.80m & 75kg, 2.20m & 90kg) for the posed body meshes since pictorial humans may have distorted proportions. Next, we determine 3D pose points of the mesh by retrieving the bone heads from the skeleton. In total, we extract 20 pose points (i.e., head, neck, thorax, pelvis, l/r shoulder, l/r elbow, l/r wrist, l/r hand, l/r hip, l/r knee, l/r ankle, l/r foot) and take the midpoint of two other pose points (i.e., l/r eye).

As a further processing step, we split the 3D body mesh into sub meshes for different body parts (i.e., head, torso, upper arms, lower arms, hands, upper legs, lower legs, feet). For this, we first iterate through the mesh vertices and derive a body part index from the maximum weight associated to each vertex group. Secondly, we iterate through the mesh triangles and assign them the same body part index as the majority of vertices of a triangle. Triangles with the same body part index are then selected and separated from the mesh. To smoothen the spikes at the boundaries, we split the two edges of a boundary triangle at their center points and assign the resulting smaller triangle to the other body part. Finally, we calculate center points for the vertices lying at any boundaries and connect them to close any arisen holes. The individual body parts are exported in OBJ format and converted to SDFs by the mesh_to_sdf[1] library.

After subdividing the body parts into watertight meshes, we create binary mask images for each body part and categorical mask images for all body parts using vertex colors and custom shaders in the rendering pipeline of Blender. Additionally, images with depth values and UV coordinates are generated using ray casting. By repositioning the orthographic camera, four views (i.e., front, left, back, right) are produced for each of the three types of 2D images.

2D body part masks, depth and UV images as well as 3D pose points and SDFs of individual body parts will serve as training and validation data for our networks. For enhancing the textures, we cartoonized heads of about 3100 humans (X. Wang & Yu, 2020) from the PASCAL-Part dataset (Chen et al., 2014). All data items are normalized to equal sizes, but their original size is stored as metadata. As testing data, we annotated 2D skeletons and body part masks of 12 larger figures from historic and contemporary pictorial maps, which mainly originate from a pictorial map classification dataset (Schnürer, Sieber, et al., 2021). The selected test figures vary in poses, clothes, genders, skin colors, drawing styles and viewing perspectives.

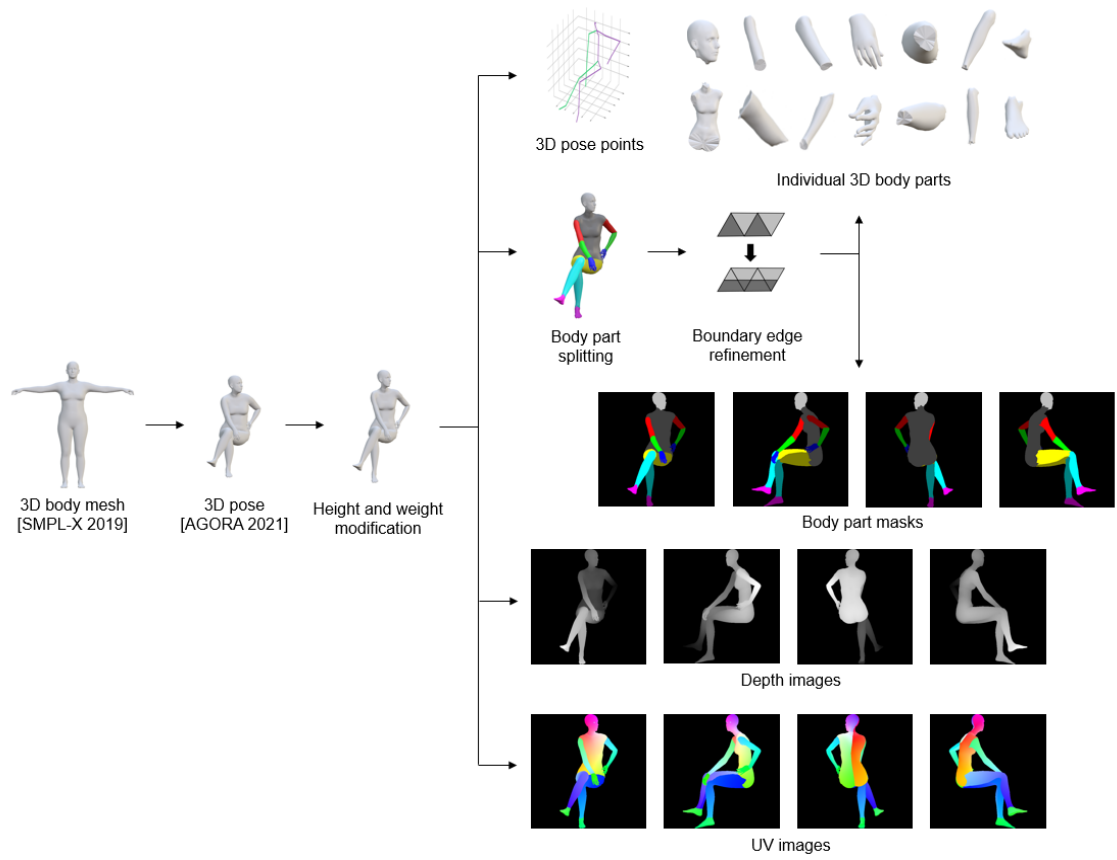

**Figure 1.** Our data processing workflow

## 4. Methods

### *4.1. 3D pose estimation*

We use a network proposed by Martinez et al. (2017) to predict depth coordinates for a 2D skeleton. The network consists only of two blocks of fully connected and dropout layers as well as a residual connection. In the original work, humans are captured by four cameras having a perspective projection. We adapt the network by introducing an orthographic projection, where the depth coordinate of the 3D skeletons is omitted to construct the projected 2D coordinates for our training data. Since pictorial figures are mostly hand-drawn in arbitrary projections and an additional network would have to be trained to estimate the parameters of a perspective camera (e.g., focal length, distortion coefficients), we apply only the orthographic projection to our test data. Nevertheless, we report the results of a hypothetical perspective camera for our validation data. Another minor modification of the original network is the addition of five skeleton keypoints – one at each hand and foot, and one at the head. Those will be helpful for the 3D body part inference in the next step.

We train each network configuration for 100 epochs using the given hyperparameters (i.e., batch size = 64, learning rate = 0.001, dropout = 0.5, batch normalization) by Martinez et al. (2017) since the authors already tested those extensively in their work. All experiments in this paper are conducted on an NVIDIA GeForce GTX 1080 and our custom architectures are implemented with TensorFlow[2]. We set the height of our figures uniformly to 1.79m, since this is the average height of the two test subjects in the original dataset according to their body meshes. Quantitative results (Table 1) show that the root mean squared errors (RMSE) of the orthographic projection are only slightly higher compared to the perspective one, whereas the percentages of correct keypoints ($PCK_{150mm}$) are slightly lower. An extreme outlier occurs, especially for the orthographic projection, when our validation data is predicted by the network trained on their data, demonstrating that re-training is necessary. After doing so, a similar accuracy to their original training and validation data is reached. The error slightly increases when adding the five pose points (i.e., l/r hand, l/r foot, midpoint between the eyes) in our skeletons. Qualitative results show that poses can be well-recovered for pictorial figures (Figure 2). Only in one out of the 12 test figures, an arm was positioned in front of the body instead of behind it (Figure 4, lower row).

| Training data | Validation data | Projection | Pose points | RMSE [mm] | $PCK_{150mm}$ [%] |
|---|---|---|---|---|---|
| theirs | theirs | perspective | 16 | 44.40 | 97.56 |
| theirs | theirs | orthographic | 16 | 45.68 | 96.59 |
| theirs | ours | perspective | 16 | 193.82 | 45.84 |
| theirs | ours | orthographic | 16 | 808.38 | 7.03 |
| ours | ours | perspective | 16 | 42.23 | 96.83 |
| ours | ours | orthographic | 16 | 46.68 | 95.10 |
| ours | ours | perspective | 21 | 48.82 | 95.63 |
| ours | ours | orthographic | 21 | 54.85 | 93.30 |

**Table 1**. Average root mean squared errors and percentages of correct keypoints of pose points for estimating depth coordinates of human poses using their (Martinez et al., 2017) and our data as well as different projections and number of pose points

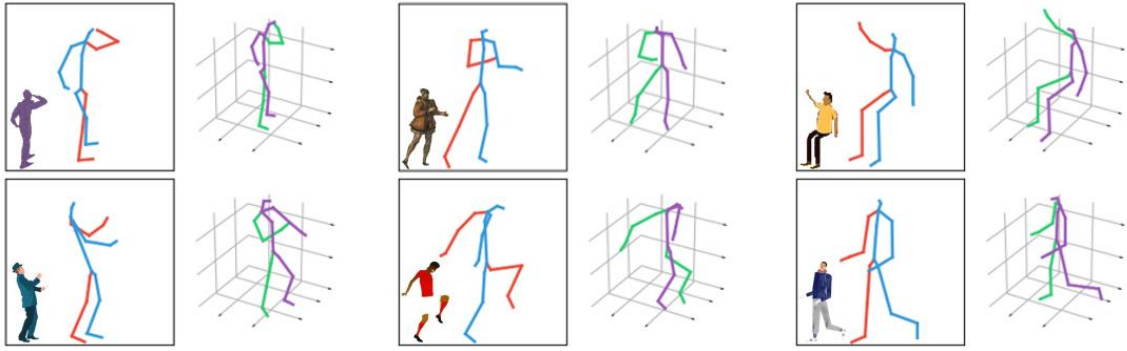

**Figure 2**. Estimated 3D poses (green/violet) from 2D poses (red/blue) of pictorial figures from our test data after training the network of Martinez et al. (2017) with our data (i.e., 21 pose points, orthographic projection)

*4.2. 3D body part inference*

We generate 3D body parts from their 2D masks by a network called DISN (W. Wang et al., 2019). Originally, the network predicts 3D SDFs of objects in 2D images, which are encoded in a series of convolution and down-sampling operations to a final feature map (i.e., global features). Intermediate feature maps of the encoding process are up-sampled and concatenated so that local features can be retrieved point-wise. Wang et al. (2019) propose two variations for the decoding part: On the one hand (i.e., DISN one-stream), the encoded 3D query points, global and local features are combined and decoded by fully-connected layers. On the other hand (i.e., DISN two-stream), 3D query points and global features as well as 3D query points and local features are first combined and decoded by fully-connected layers in parallel, and finally added.

      We extend the network by additionally concatenating 3D pose points, which have been estimated in the previous stage, to the global and local features together with the 3D query points. We use two 3D pose points as anchor points for each body part (e.g., elbow and wrist for a lower arm), except for the torso where four points are used (i.e., left/right hip, left/right shoulder). The pose points embed information about

orientation and depth of the body parts. This has the advantage that an initial network proposed by Wang et al. (2019) can be omitted, which estimates translation and rotation parameters to transform points from world space into camera space. Since we feed only 64x64px masks into the adapted network, we reduced its parameters (Appendix A). We sample the same amount of positive and negative SDF values (i.e., 2000 each) to get a distinct zero-iso-surface. The distance values are sampled randomly within the cubic grid to recover coarse structures and fine details of body parts, however leading to an accuracy trade-off in either granularity.

We report errors for inferring body parts with and without pose points for the one-stream and two-stream architecture for hands (Table 2). We selected a hand as exemplary body part for our measurements since fingers are the most difficult structure to recover (Table 3). We train each configuration five times for 200 epochs at a learning rate of 0.0001 and a batch size of four. Results show that errors are similar for the two architectures, and decrease with the additional pose points in both cases. To construct all body parts (Figure 4), we mirror symmetric body parts (e.g., right and left foot).

|         | **DISN one-stream** | | **DISN two-stream** | |
|---------|---------------------|------------------|---------------------|------------------|
|         | without pose points | with pose points | without pose points | with pose points |
| **RMSE** | 0.063 | 0.051 | 0.063 | 0.050 |
| **IoU [%]** | 47.80 | 52.63 | 47.79 | 53.31 |

**Table 2**. Average root mean squared errors and intersections over unions on our validation data for inferring 3D SDFs of hands from 2D masks

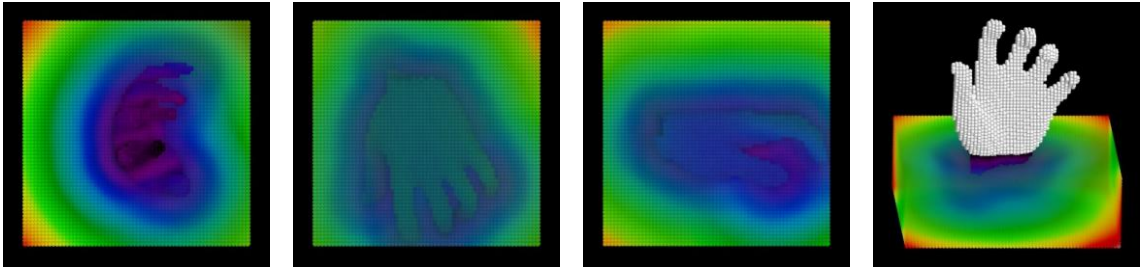

**Figure 3:** SDF around a hand viewed from the top, back and left, and in oblique perspective. Only positive distance values (blue = small, green = intermediate, red = large distances) are shown. A 3D segment is depicted in oblique perspective.

|         | Torso | Head  | Upper arm | Lower arm | Hand  | Upper leg | Lower leg | Foot  |
|---------|-------|-------|-----------|-----------|-------|-----------|-----------|-------|
| **RMSE**    | 0.014 | 0.011 | 0.024     | 0.029     | 0.050 | 0.015     | 0.013     | 0.019 |
| **IoU [%]** | 87.01 | 93.01 | 80.42     | 74.23     | 53.31 | 87.65     | 83.66     | 81.19 |

**Table 3**. Average root mean squared errors and intersections over unions on our validation data for inferring 3D SDFs of different body parts from 2D masks using DISN two-stream with pose points

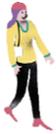

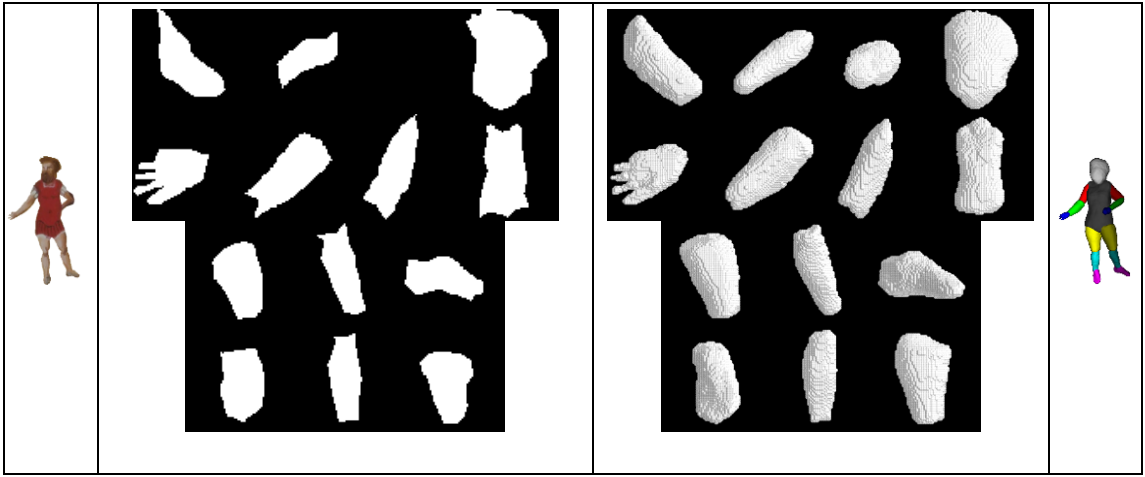

**Figure 4**. Inferred body part SDFs (distance < 0) from masks by the DISN one-stream network with concatenated pose points. Each network producing a 3D body part is trained individually. The mask of the left hand of the second figure is empty since it is hidden in the original image.

*4.3. UV coordinates prediction*

We predict UV coordinates, ranged zero to one, from a depth image and body part masks of the figures by designing a network similar to U-Net (Ronneberger et al., 2015). The input data is derived from the outputs of the previous two steps. Each 3D body part is positioned at the mid points of the bones of the 3D skeleton to form the full body. The size of each body part is determined by the height and width of the 2D body part mask as well as enclosed 3D skeleton points. For the latter, a multi-layer perceptron - consisting of three layers with 20, 40, 20 neurons - predicts the size from the enclosed 3D skeleton coordinates to compensate the lack of depth information of the 2D body mask.

By assembling the inferred 3D body parts (Appendix B), we derive depth images and body part masks for four camera views (i.e., front, left, back, right). The additional front view will be helpful to generate textures for overlapping parts later-on. Since the projection of the drawn figures may vary, we simply assume an orthographic projection. We feed depth images and body part masks in batches of eight and resized to

256x256px into our U-Net-like network (Figure 5). Our network consists of 1- and 2-strided convolutions, which are used for down- and up-sampling, as well as skip connections. The network is trained for 50 epochs at a learning rate of 0.001 and for another 50 epochs at a learning rate of 0.0001 with the Adam optimizer. Since the loss converged at similar values, we report the results of one single run. It turned out that the additional body part masks, which are multiplied with the depth image, lead to a lower error compared to training with depth images only for our validation data (Table 4). Smooth UV images were produced for our test data (Figure 6).

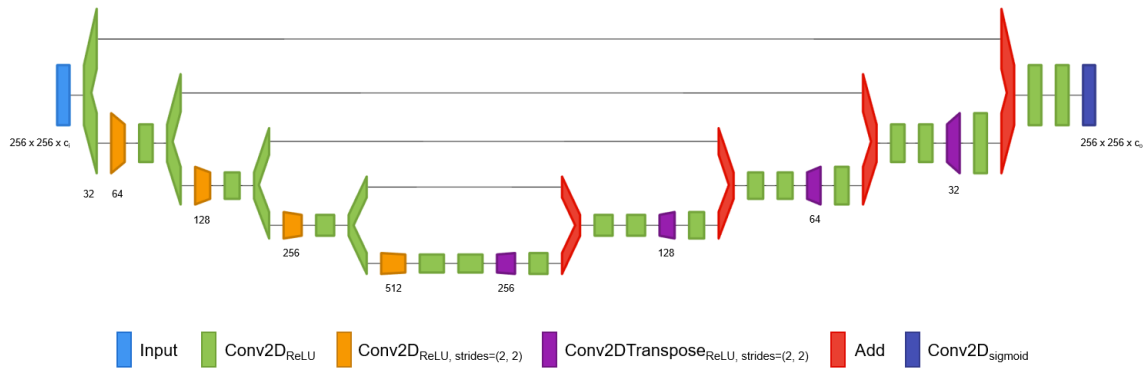

**Figure 5**. Network architecture for predicting UV coordinates from a depth map. $c_i = 1$ for inputting a depth image; $c_i = 14$ for inputting a depth image multiplied by body part masks; $c_o = 3$ for outputting the two UV coordinate channels and a body mask channel used in the loss calculation. Numbers below each layer denote the channel dimension. Figure created by Net2Vis (Bäuerle et al., 2021).

| Input data | UV coordinates (MAE) | UV coordinates (RMSE) |
|---|---|---|
| Depth image | 0.014 | 0.042 |
| Depth image * Body part masks | 0.011 | 0.031 |

**Table 4**: Mean absolute errors (MAE) and root mean-squared errors (RMSE) for predicting UV coordinates of pictorial human figures from different input data

| | Input depth image and body part masks | Output UV image |
|---|---|---|
| 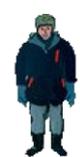 | 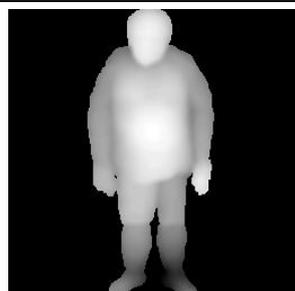 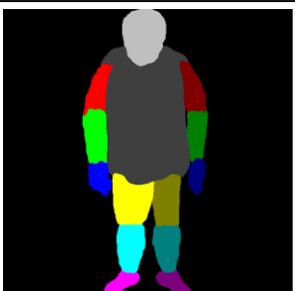 | 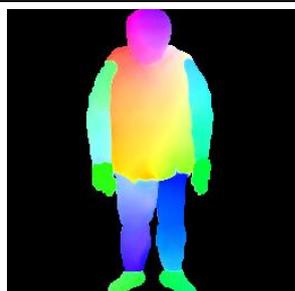 |
| 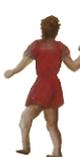 | 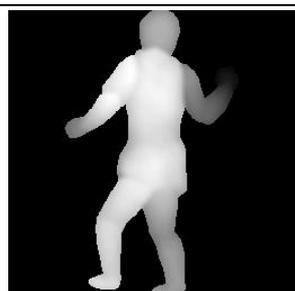 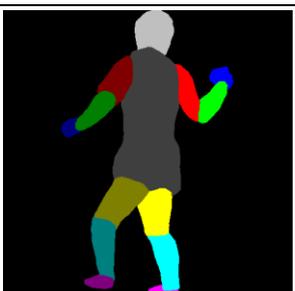 | 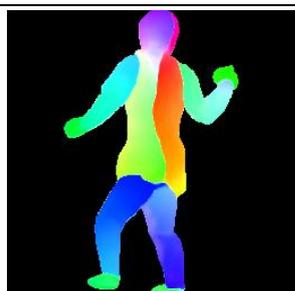 |

**Figure 6**. Predicted UV coordinates of pictorial human figures from a depth image and body part masks by our fully convolutional network. The body part masks are one-hot encoded in 14 channels for the neural network. The UV coordinates are stored in two channels and have been mapped to a squared color circle (Figure 8) for visualization purposes.

## *4.4. Texture inpainting and enhancement*

We create 256x256px textures when viewing the figure from behind, the left and the right by a generative network (Grigorev et al., 2019). Due to minor mismatches of the shape between the predicted UV coordinates in the previous step and the given texture, we input the intersection of both into the network. We add a grey rectangle to the background since the network was trained on human models standing in front of a white wall, which appears greyish due to lighting and shadows. As a post-processing step, we crop the output images to the given body masks.

Since the authors did not publish the code for training their generative network, we could only use a pre-trained version and thus report qualitative results only (Figure 7). We apply texture maps, whose color values were retrieved from the inpainted textures using the predicted UV images (Figure 8), to the body parts to render the final images (Appendix C). In general, colors of clothes, skin and hair were mostly plausibly generated. Artifacts appear for uncommon poses (e.g., sitting, football playing) and at the shoes/feet. Coarse texture structures could be created, however pictorial black strokes representing folds in the textures were not transferred from the source image. To take these pictorial characteristics into account, a cartoonization network (X. Wang & Yu, 2020) can be applied to the inpainted textures.

Since some of the inpainted faces clearly originate from real humans, we train an autoencoder to map them to a more pictorial style and to possibly recover missing facial details. We establish a shallow branch (Gondara, 2016) for denoising the hue and saturation channel, and a deeper branch including a bottleneck for learning structures and shadings in the value channel (Figure 9). Colors (i.e., hue and saturation) and shadings (i.e., value) are weighted equally in the loss function. We augment the realistic input images by blurring and oilpainting, by adding noise (i.e., gaussian, salt and pepper) and by varying the hue. The target images have been converted from the input images by the above cartoonization network. The autoencoder is trained for 200 epochs at a batch size of 32 and at a learning rate of 0.001 with the Adam optimizer. During training, head images with varying looks can be obtained (Figure 10). Convincingly painted results, however, are rather the exception (i.e., roughly 5% of the test images).

| Input texture and UV maps | Inpainted texture | Cartoonized texture |
|---|---|---|
| 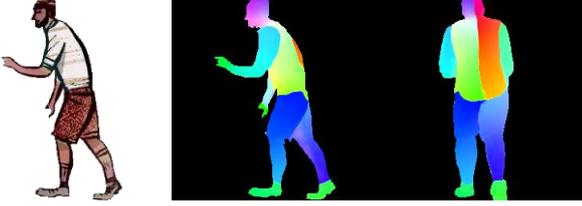 | 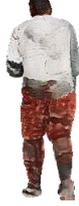 | 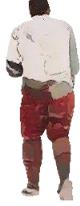 |
| 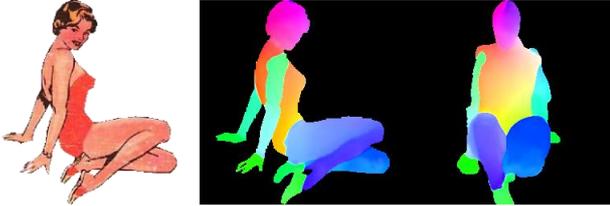 | 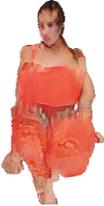 | 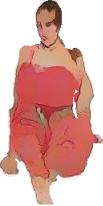 |

**Figure 7.** Generated textures (already cropped to the input mask) of two pictorial figures from a given image as well as source and target UV maps by the inpainting network. The inpainted image is cartoonized to mimic a more pictorial style.

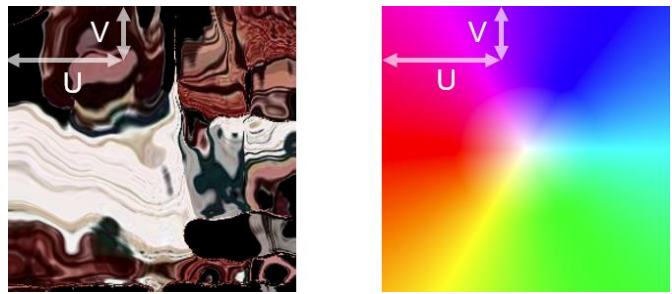

**Figure 8.** UV coordinates in the inpainted texture (left) and in the color wheel (right) at the position of the left eye of a pictorial figure (Figure 7, upper row)

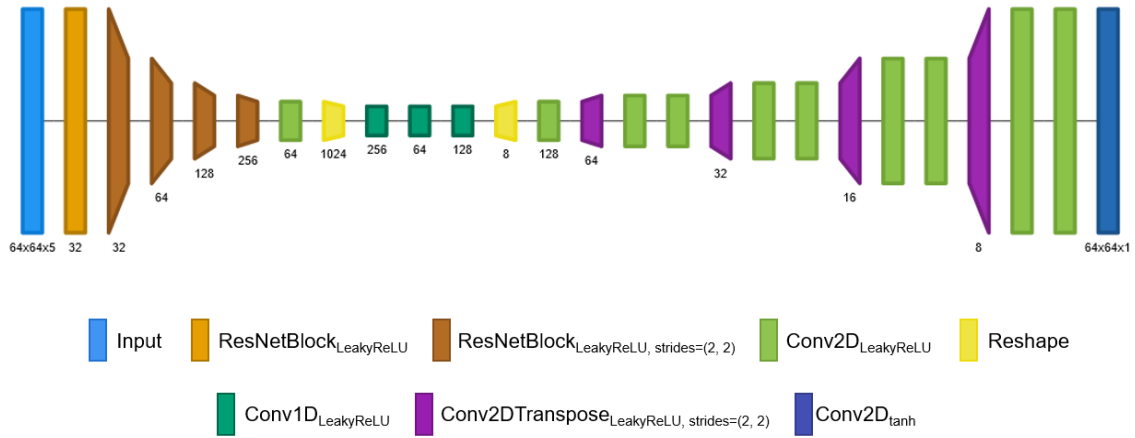

**Figure 9.** Network architecture for enhancing textures of inpainted heads. We input HSV images together with predicted UV-coordinates to account for the different orientations of the heads. This branch outputs the value of the color, whereas hue and saturation are outputted by another branch. Figure created by Net2Vis (Bäuerle et al., 2021).

| **Input textures and UV maps** | **Enhanced textures** |
|---|---|
| | |
| | |
| | |
| | |

**Figure 10.** Denoised and resynthesized textures of inpainted heads from different views (i.e., right, front, left, back)

*4.5 Real-time rendering*

The inferred figures can be rendered in real-time using the sphere tracing algorithm (Hart, 1996). A 256x256px image is rendered at 25 frames per second (FPS) and a 512x512px image at 11 FPS with an NVIDIA GeForce 1080 GTX even when enabling computationally intensive features (i.e., trilinear interpolation, normal calculation, texture blending). Optionally, we can enable a perspective projection by sending rays from a point location, however differences to the orthographic projection are only marginal. Also, diffuse lighting can be added by calculating the angle between a virtual light source and the surface normals.

To integrate the figures into existing 3D map environments such as virtual globes, which are mainly based on the traditional rendering pipeline, they can be rendered with transparent background in billboards, while the sphere tracing algorithm is implemented in the fragment shader (Schnürer et al., 2017). Another option is to export a point cloud by returning the 3D coordinates of surface intersections in the ray marcher. The point cloud can be further turned into a triangle mesh by the ball-pivoting algorithm (Bernardini et al., 1999). We exemplarily illustrated the outcome of the latter two conversion steps by placing the 3D figures on the original map in a 3D modelling software (Figure 11) and a virtual globe toolkit (Figure 12).

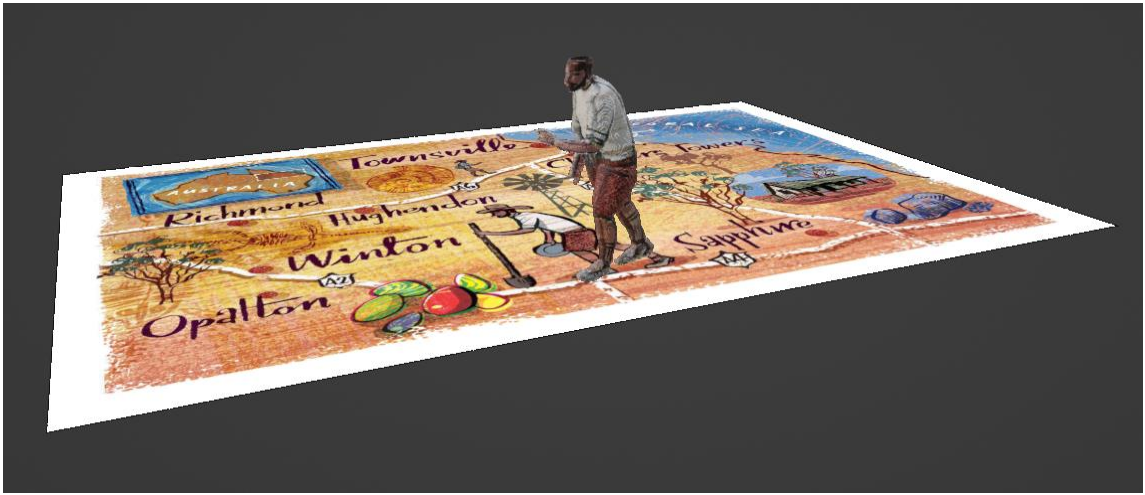

**Figure 11.** Remeshed pictorial 3D figure placed on the original map[3] in Blender[4]

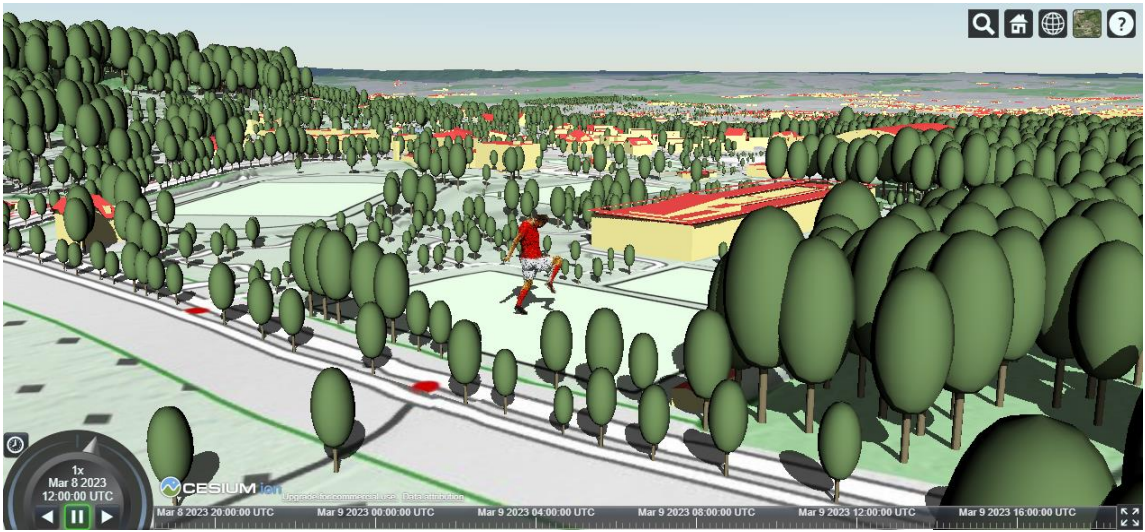

**Figure 12.** Remeshed pictorial 3D figure in CesiumJS[5]. The figure is placed in front of the FIFA headquarters, similarly to the original map[6]. 2D base map, 3D buildings and trees originate from swisstopo's GeoAdmin API[7].

## 5. Use case

We see a large potential in adding the constructed figures as protagonists or secondary characters to story maps. Particularly, 3D maps convey the topography vividly and allow channeling the depicted topic by occlusions (e.g., mountains, trees, fog). Instead of presenting multimedia content in overlays (e.g., Matt, 2019), we suggest placing essential figures or other objects directly and in a consistent style on the map to support story.

In the following, we outline how a story map including characters may be designed for children (Figure 13). We take Charles Darwin's (1839) round-the-world journey as an example, specifically a stop in Port St. Julian, Patagonia, in January 1834. After a short introduction to this setting, the user can visit different places in any order. We provide different incentives to follow the story, and to interact with characters and the environment (Table 4). The story ends after having explored all places.

Pedagogically, our proposed story map may improve map and visualization literacy (e.g., route planning, interpretation of climate diagrams), and may help to correct misconceptions (e.g., Patagonians perceived as giants). It would connect interdisciplinary fields, such as biology (e.g., penguins), history (e.g., early explorers), and ethnology (e.g., indigenous people). Although some gamification elements are included, it is intended to put the focus on the scientific aspects. Our presented pipeline of different machine learning models will help map creators in constructing and potentially animating 3D human figures.

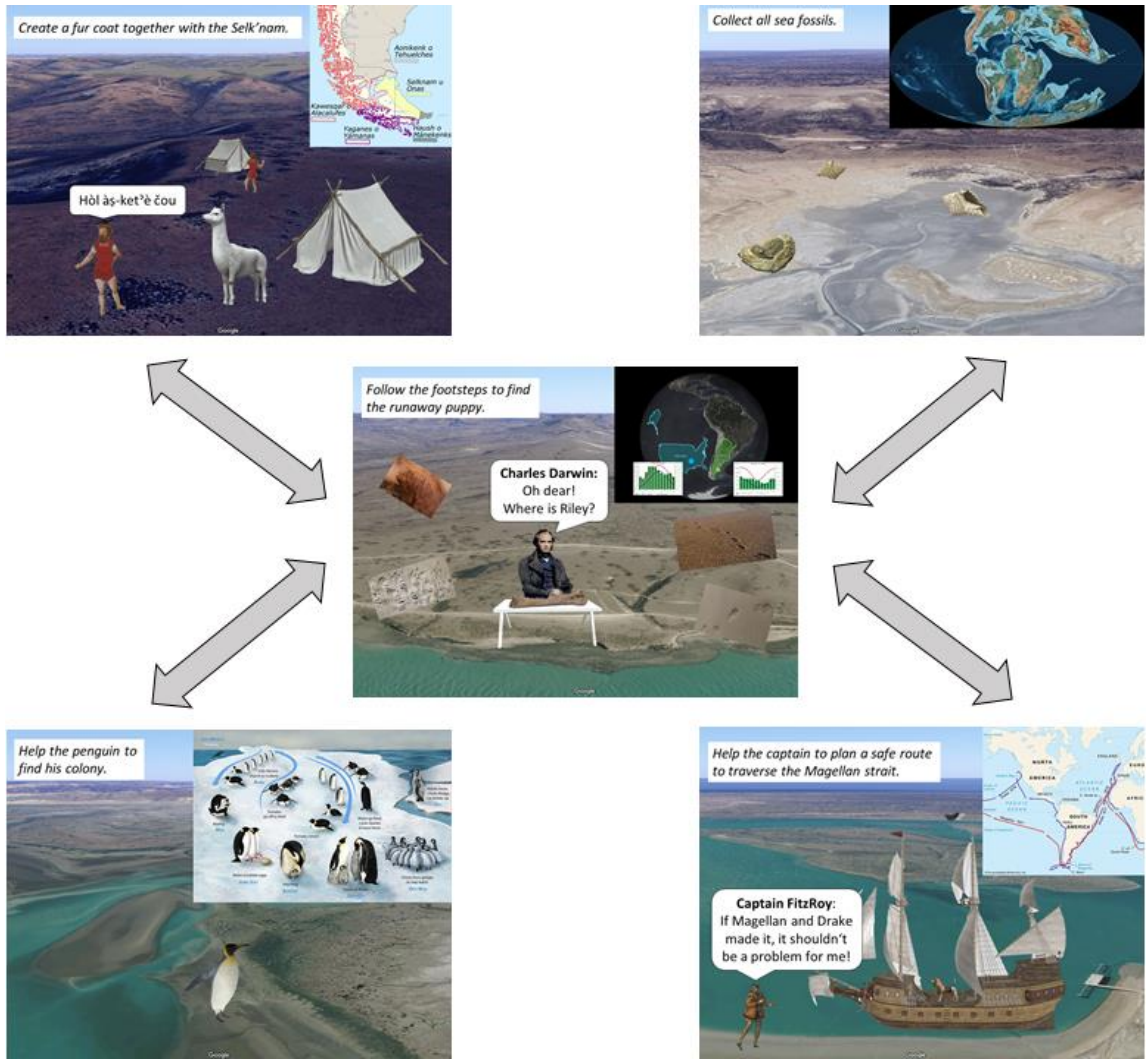

**Figure 13**: Sketched story map for children about Charles Darwin in Patagonia (sources: Appendix D). The start location is in the middle and possible places to explore are at the corners. At each place, a puzzle needs to be solved after being provided with some contextual information (top right graphic of each scene). The reconstructed figures will be part of the story.

| | |
|---|---|
| **Attraction** | - having an overarching goal (e.g., finding a runaway dog)<br>- curiosity to explore the remote area (e.g., by following footsteps) |
| **Affection** | - smooth transitions between local and global scale (e.g., by giving background information to the current scene)<br>- temporal animations (e.g., plate tectonics) |
| **Interaction** | - personal stories of characters (e.g., the captain)<br>- small tasks to fulfil (e.g., assembling a piece of clothing) |
| **Comeback** | - variations in tasks and questions (e.g., by randomization)<br>- a different ending (e.g., the runaway dog reappears at different locations) |

**Table 4**: Application of storytelling concepts (Thöny et al., 2018) to our sketched story map

## 6. Discussion

Our work is a continuation of another computer vision experiment for pictorial maps (Schnürer, Öztireli, et al., 2021), where silhouettes of figures, their body parts and pose points have been segmented by two neural networks. Therefore, it can be assumed that these data items can be automatically extracted. Nevertheless, we annotated our test data manually to have a solid foundation for the current experiment. For our training data, we varied the sizes and weights of body meshes of real humans, which had a positive impact on reconstructing a test figure with thin long limbs and a small head. The consideration of clothes would definitely improve the quality of reconstruction, for instance, our current pipeline does not support hats. However, publicly available training datasets of clothed humans did not contain 3D body meshes or body part segmentations. Beyond, some clothes (e.g., long skirts) would behave differently than the underlying body parts, but we aimed primarily at skeletal animation as a follow-up use case.

   Besides selecting and enhancing the training data, the structure of the different networks may be modified. We showed that increasing the number of joints from 16 to 21 led to a 1cm higher error for the 3D pose estimation network, which is still tolerable and provided a benefit to the reconstruction of hands, feet and the head. We used captured poses of mainly standing persons as training data; alternatively bones of a skeleton may be oriented according to a range of possible joint angles (Soucie et al., 2011) to better handle uncommon poses. We did not consider including joints of fingers, which would have been available in the SMPL-X training dataset, since hands of pictorial figures are usually small, their fingers are not easily distinguishable and sometimes contain less than five fingers. Instead of estimating the 3D coordinates directly, relative rotation angles of joints or limbs may be predicted, however this would require more complex and constrained network designs, as noted by Martinez et al. (2017). Estimating more camera parameters would probably increase the prediction accuracy, yet we achieved satisfactory results by simply assuming an orthographic projection, similarly to other works (e.g., Huang et al., 2020).

While no fine-tuning was necessary for the pose estimation network, we carefully configured DISN to infer body parts. We also examined normalizing the data (e.g., sqrt/log transform), using 3D deconvolutions (instead of query points), sampling points near the surface (instead of equally spaced grid points), predicting a top and side view (additionally to the front view), outputting a binary field (additionally to the SDF); but those did not significantly improve the reconstruction quality. Generally, silhouette-based 3D reconstruction is a more difficult task than a texture-based recovery since textures may contain depth information, thus the qualitative results are only partly comparable to those of the original network. The addition of pose points helped to recover partly or totally hidden body parts, though not visible hands are approximated by ellipsoids, which is the average shape. Other issues concern the rather realistic forms of the body parts and gaps between them (Figure 14).

Predicting the UV coordinates from the depth map was a straight-forward task by using a fully convolutional network similar to U-Net (Ronneberger et al., 2015). Inputting depth maps only would have been already sufficient to get a smooth image for the validation data (i.e., real humans). However, on our test data (i.e., pictorial humans), where the depth map is derived from the inferred 3D body parts, stains appeared on the UV image, which could be remedied by additionally feeding the masks of the body parts into U-Net. We refrained from predicting UV coordinates or even texture colors together with the body parts because "color prediction is a non-trivial task as RGB colors are defined only on the surface while the [signed distance] field is defined over the entire 3D space" (Saito et al., 2019, p. 2308).

Although the inpainting network is biased on generating textures of real humans, it produced adequate results on a coarse level for pictorial humans. For better matching the texture with the input, the potential of symmetries could be exploited (Zhou et al., 2021). Furthermore, texture maps probably need to be deformed (Shu et al., 2018) in case the body part shapes deviate much. Since texture mappings for occluded body parts may be incorrect when viewing the figure from another than the four perspectives, one may increase the number of textures to prevent these artifacts. Due to the lack of adequate training data, we cartoonized photos of humans to train our autoencoder, however the resulting textures are still quite realistic. Our autoencoder is able to recover some facial details, yet a more expressive latent space may be learned by a variational autoencoder. Overall, an end-to-end network would be desirable instead of our four consecutive networks to benefit from synergy effects.

To realize our exemplary use case, existing story map editors (e.g., ArcGIS StoryMaps[8]) would need to be extended to support different storylines, game templates, and textual options. The availability of a 3D model store would facilitate the reusability and the copyright management of the reconstructed figures and other objects. Positioning and viewing perspective, animations parameters and triggers, interactions with the map content and other characters, and possibly cartographic functions may be defined for the characters in the story map editor. In cognitive experiments, optimal parameters (e.g., animation speed) would need to be clarified in detail and whether the approach including figures in story maps generally offers an added value to the user.

| Problem | Visual example | Possible solution |
|---|---|---|
| Gaps may appear between body parts | 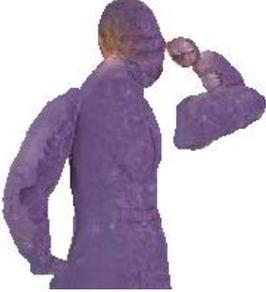 | Refine transformation parameters (i.e., translation, rotation, scaling) of body parts by optimization |
| Non-visible hands are approximated by ellipsoids | 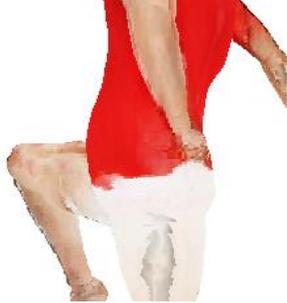 | Adapt a hand template based on the lower arm thickness or by parameters of the visible hand |
| More abstract geometric shapes are not supported | 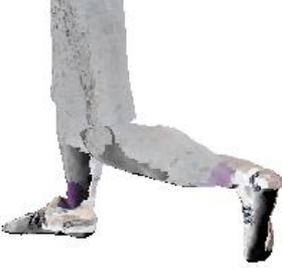 | Add synthetic training data (e.g., cuboids, ellipsoids) and learn a 3D deformation field, which uses body parts of real humans as templates |

**Figure 14**. Failure cases of inferred body parts

## 7. Summary and future work

In this work, we generated implicit 3D representations of human figures appearing on pictorial maps using machine learning methods (Figure 15). We showed that plausible poses and body parts can be inferred when training the networks with data of real humans. However, we see a need for improvement in refining shapes and textures, in supporting hair and clothes as well as in simplifying the workflow. Our automated workflow runs only a few minutes, whereas manual sculpting and texturing of the 3D models would take several hours.

Next to human figures, also other pictorial entities like animals, sea monsters or ships may be transferred to the third dimension. Moreover, 3D buildings or distinct landscape features may be derived from historic maps in oblique view (e.g., Murer, 1576). Cartography would be rather concerned with abstract and georeferenced representations while the reconstruction of detailed representations would rather fall into the domains of other fields (e.g., archaeology, anthropology, evolution biology).

The automatic 3D reconstruction of buildings and landscape will accelerate the development of 'time travel' applications, where users can see the past and future of a geographic area (e.g., Stadt Zürich, 2022). When additionally viewing a 3D city in virtual reality, users will get a more vivid impression of its anatomy (e.g., the narrowness of alleys). As envisioned in the concept of the metaverse, the virtual space may be populated with avatars, where our animation-ready 3D pictorial human figures come into operation. We would propose the term 'cartoverse' for this kind of cartographic metaverse. Several challenges would need to be addressed in the future to provide a fully functional cartoverse, for instance, how to avoid motion sickness during spatial navigation or for different perspectives, how to interact with 3D objects via speech or gesture recognition, or how to present thematic information additionally to the topographic elements.

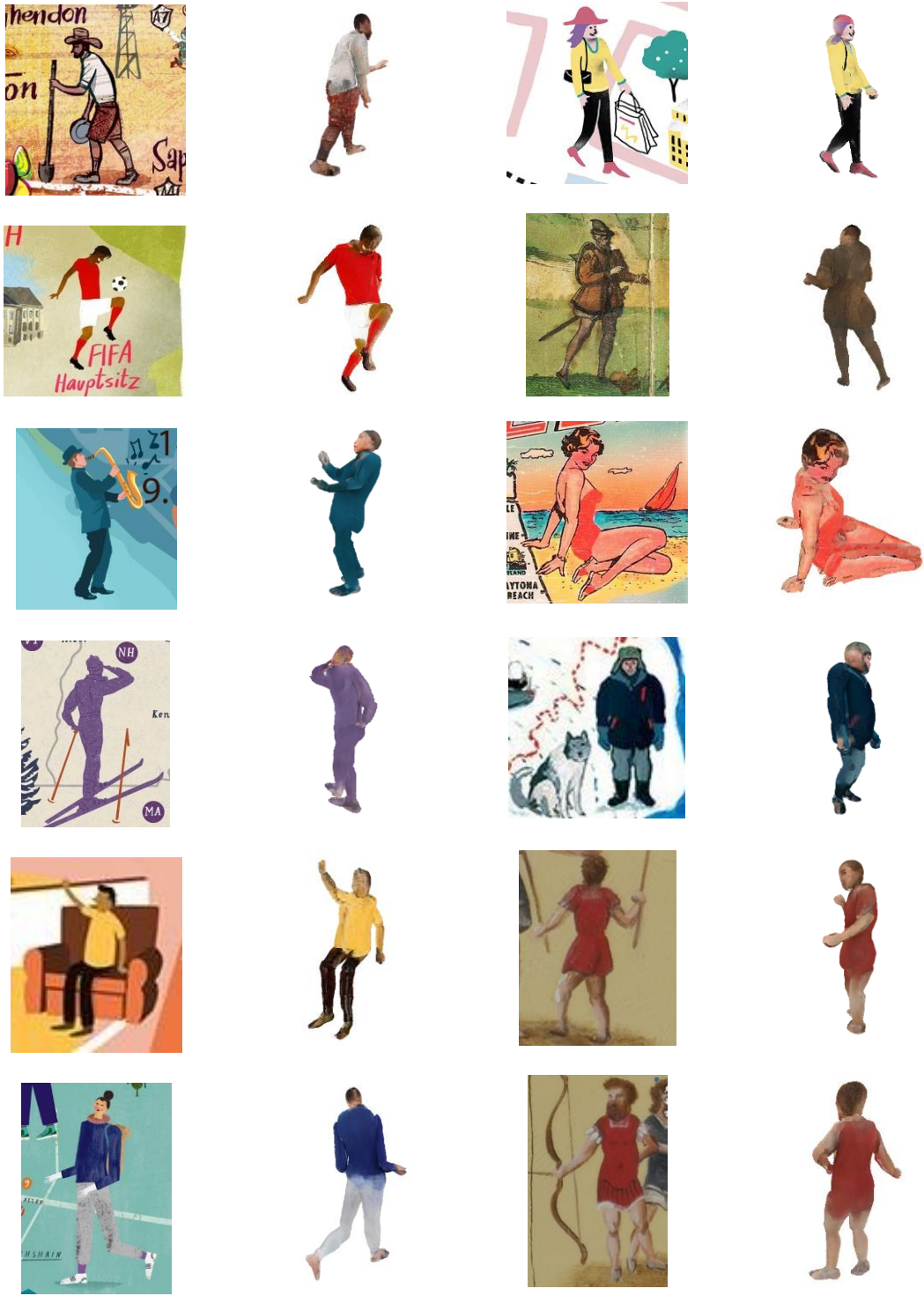

**Figure 15.** All pictorial human figures on maps from our test dataset and our inferred 3D models from different views

# Notes

This article extends the abstract "Inferring Implicit 3D Representations from Human Figures on Pictorial Maps" submitted to the International Cartographic Conference in Florence 2021.

# Footnotes

1. https://github.com/marian42/mesh_to_sdf
2. https://www.tensorflow.org/
3. https://www.behance.net/gallery/30454283/Queensland-National-Geographics-Traveller-Mag/modules/195508163
4. https://www.blender.org/
5. https://cesium.com/
6. https://i.pinimg.com/originals/8e/e9/7e/8ee97e45c6d6fd64a96a6bff08144719.jpg
7. https://api3.geo.admin.ch/services/sdiservices.html
8. https://storymaps.arcgis.com/

# Acknowledgements

We like to thank Jost Schmid-Lanter and his colleagues from the map department of the Zurich Central Library for providing images of figures from a digital replica of the St Gallen Globe.

# Funding

This work was supported by a UKRI Future Leaders Fellowship [grant number G104084].

# Disclosure statement

No potential conflict of interest was reported by the authors.

# Data availability statement

Training and test datasets, models, and source code are available at: http://narrat3d.ethz.ch/

# Appendix

## A. Adaptations to the Deep Implicit Surface Network (DISN) for 3D body part inference

Query points and additional pose points, which are inputted into the network, are encoded by three 1D convolutions with four times smaller filter sizes (i.e., 16, 64, 128). In the image encoder, we reduced the number of filters for convolutions by four (i.e., 8, 16, 32, 64, 128) and use only one convolution layer after a strided convolution layer at each level. Similarly, we lowered the size of the global features by four (i.e., 256). The filter sizes for 1D convolutions in the decoder part remained unchanged (i.e., 512, 256, 1).

## B. Derivation of depth maps and body part masks for UV coordinates prediction

We implemented a custom ray marcher in Python using the library 'Numba', which enables executing parallel operations on the GPU, for rendering the inferred 3D SDFs. An SDF value denotes the shortest distance (d) to object surfaces, while the sign indicates whether a point lies inside ($d < 0$), on ($d = 0$) or outside ($d > 0$) the object. We trilinearly interpolate the eight closest grid points of the evenly spaced SDF to get smoother surfaces. The SDFs of the different body parts can be combined by the union operation (i.e., $\min(d_1, d_2)$). Afterwards, a depth image can be obtained by iteratively cumulating the covered distances from the virtual camera to the body surface during each ray marching step. As an enhancement, we smooth the depth map by a 5x5 averaging filter. A body part mask is produced by returning the index of the first hit body part along the ray.

## C. Application of textures to figures

The ray marcher (Appendix B) is extended by projecting the inpainted texture maps to the surfaces of the 3D body parts from four views (i.e., front, back, left, right). As an enhancement, we blend the obtained textures, what means that the steeper the angle of the surface normal to the texture, the more weight the color value gains from this texture. The normals are approximated by nearby SDF values at the surface points (i.e., n = [SDF(x+ε, y, z) - SDF(x-ε, y, z), SDF(x, y+ε, z) - SDF(x, y-ε, z), SDF(x, y, z+ε) - SDF(x, y, z-ε)]). For an interactive view, we pass the rendered images to the canvas of the library 'matplotlib', where mouse events can be captured to zoom and rotate around the depicted figure.

## D. Sources of the 3D story map with pictorial figures

| **Maps** | |
|---|---|
| Background maps | https://www.google.com/maps |
| Globe for comparing sizes of countries | https://arnofiva.github.io/world-sizes/ |
| Natives in Patagonia | https://commons.wikimedia.org/wiki/File:Pueblos_ind%C3%ADgenas_de_la_Patagonia_Austral.svg |
| Lifecycle of penguins | https://de.wikipedia.org/wiki/Datei:Penguin-lifecycle-de.svg#/media/Datei:PENGUIN_LIFECYCLE_H.JPG |
| Plate tectonics | Scotese, C.R. (2016). Plate Tectonics, Paleogeography, and Ice Ages, (Modern World - 540Ma), YouTube Animation. Available at: https://youtu.be/g_iEWvtKcuQ |
| Seafarers' voyages | https://www.britannica.com/biography/Ferdinand-Magellan/Circumnavigation-of-the-globe |
| | |
| **Images** | |
| Charles Darwin | https://commons.wikimedia.org/wiki/File:Charles_Darwin_by_G._Richmond.jpg |
| Penguin footprints | https://commons.wikimedia.org/wiki/File:Penguin_footprints_on_the_beach_%285565682274%29.jpg |
| Human footprints | https://commons.wikimedia.org/wiki/File:Punta_Prosciutto_footsteps.jpg |
| Large footprints | https://commons.wikimedia.org/wiki/File:Nodosaur_Footprint_Verified_-_Detail_of_Baby_Footprint_(7846740914).jpg |
| Small footprints | https://commons.wikimedia.org/wiki/File:Footsteps_on_the_beach,_Seaford_-_geograph.org.uk_-_2599216.jpg |
| Climate charts | https://www.climatestotravel.com/ |
| | |
| **3D models** | |
| Bone | https://sketchfab.com/3d-models/horse-bone-d51e7216a5bb41bea3ee2576fa92eedc |
| Table | https://sketchfab.com/3d-models/table-for-building-91d5f0058b734eafad16a3e43070ebe6 |
| Alpaca | https://sketchfab.com/3d-models/alpaca-non-commercial-5de8754563254e34837ec4aacac8632e |
| Tent | https://sketchfab.com/3d-models/tent-fa46028e8d3849399ba5271df07ed99c |
| Penguin | https://sketchfab.com/3d-models/emperor-penguin-310f1d21cf534fd0bcf073aa9b08a740 |
| Fossil | https://sketchfab.com/3d-models/trilobite-fossil-b4c9b051a23445bfafcc9953deb54cc5 |
| Shell | https://sketchfab.com/3d-models/seashell-fossil-bc4b85625dd045608b498c41f8b5c1a7 |
| Sailing ship | https://sketchfab.com/3d-models/segelschiff-sailing-ship-updated-pbr-f349ecd6b81c4c25aa5d628e8913048e |
| Dog | https://sketchfab.com/3d-models/beagle-341cb7dd930a427eab3f8e925718e6c0 |
| | |
| **Miscellaneous** | |
| Selk'nam language | https://ids.clld.org/contributions/311 |